\documentclass[10pt,twocolumn,letterpaper]{article}

\usepackage{cvpr}
\usepackage{epsfig}
\usepackage{graphicx}
\usepackage{amsmath}
\usepackage{amssymb}

\usepackage{makecell}
\usepackage{multirow}

\usepackage[latin9]{inputenc}
\usepackage[english]{babel}
\usepackage{caption}
\usepackage{tabulary}

\usepackage[para]{threeparttable}
\usepackage{array,booktabs,tabularx}

\usepackage{siunitx}

\usepackage[flushleft]{threeparttablex}
\usepackage{footnote}

\usepackage[breaklinks=true,bookmarks=false]{hyperref}

\cvprfinalcopy 

\begin{document}

\title{Post-training deep neural network pruning via layer-wise calibration}

\author{Ivan Lazarevich\\
Intel Corporation\\
{\tt\small ivan.lazarevich@intel.com}
\and
Alexander Kozlov\\
Intel Corporation\\
{\tt\small alexander.kozlov@intel.com}

\and
Nikita Malinin\\
Intel Corporation\\
{\tt\small nikita.malinin@intel.com}
}

\maketitle

\begin{abstract}
We present a post-training weight pruning method for deep neural networks that achieves accuracy levels tolerable for the production setting and that is sufficiently fast to be run on commodity hardware such as desktop CPUs or edge devices. We propose a data-free extension of the approach for computer vision models based on automatically-generated synthetic fractal images. We obtain state-of-the-art results for data-free neural network pruning, with $\sim$1.5\% top@1 accuracy drop for a ResNet50 on ImageNet at 50\% sparsity rate. When using real data, we are able to get a ResNet50 model on ImageNet with 65\% sparsity rate in 8-bit precision in a post-training setting with a $\sim$1\% top@1 accuracy drop. We release the code as a part of the OpenVINO\textsuperscript{TM} Post-Training Optimization tool\footnote{\href{https://docs.openvinotoolkit.org/latest/pot_README.html}{https://docs.openvinotoolkit.org/latest/pot\_README.html}}.
\end{abstract}

\section{Introduction}

Deep neural network (DNN) models have achieved unprecedented accuracy in several crucial domains such as computer vision and natural language processing. Despite the success of DNN models, an unreasonably large amount of computations and memory required for their inference limits their deployment on edge devices, such as smart cameras equipped with low-power CPUs, GPUs or ASIC accelerators. Significant efforts in recent years have been devoted to both hardware design and algorithmic approaches to DNN model compression to enable inference speedups for various model architectures and use cases. Some of the DNN compression methods, such as 8-bit quantization, were adapted to the post-training setting where the original DNN model to be compressed could come from any software framework and no access to the original training pipeline and the training dataset is given. One of the promising approaches to reduce the memory footprint and inference latency of DNNs is weight pruning \cite{blalock2020state, gale2019state}, which results
in models with sparse weight matrices. Recently,  a lot of research and development has been aimed at leveraging weight sparsity to achieve inference speedups on a range of hardware platforms \cite{elsen2020fast, guo2020accelerating}. However, relatively little effort was devoted to providing accurate sparse DNN models in the post-training scenario.

In this work, we propose a recipe for fast post-training pruning of DNNs that produces models with significant sparsity rates (e.g. 50\%) but negligible accuracy drops. Furthermore, if combined with weight quantization techniques, the proposed method could reduce the model memory footprint by a factor of 6-8x \cite{liu2020layerwise}. We propose a fast data-free extension of our weight pruning pipeline which allows getting state-of-the-art accuracy levels for a range of computer vision models. To streamline the deployment process of sparse quantized DNNs on hardware, we have implemented the proposed method as a part of the OpenVINO\textsuperscript{TM} Post-Training Optimization tool. \\ 

We summarize our contributions as follows:
\begin{itemize}
    \item A recipe for post-training weight pruning with demonstrated results on a wide range of models and datasets. 
    \item State-of-the-art results for data-free weight pruning of computer vision models using synthetic fractal images for model compression.
    \item An ablation study of the proposed post-training weight pruning pipeline demonstrating the effects of particular components such as per-layer sparsity rate selection criteria, bias correction and layer-wise fine-tuning settings.
\end{itemize}

\begin{table*}[!ht]
\centering
\caption{Accuracy values of the sparse 8-bit quantized DNN models obtained with the proposed post-training method. Metric values were measured on a CPU with OpenVINO\textsuperscript{TM} 2021.3 as an inference engine. The same is for other accuracy values reported in the paper unless specified otherwise.}
\label{table:sparse_int8_results}
\begin{tabular*}{\linewidth}{S @{\extracolsep{\fill}}
                        *{5}{S[group-separator = {,},
                               group-minimum-digits = 6
                                ]}}
\toprule
{Model} & {Dataset (acc. metric)} & {Sparsity rate, \%} & {Compressed model acc.} & {Absolute acc. drop} \\
\hline 
{ResNet50} & {ImageNet (top@1 acc.)} & {65} & {75.09} & {1.04}  \\
{ResNet18} & {ImageNet (top@1 acc.)} & {50} & {68.93} & {0.81}  \\
{GoogleNetV4} & {ImageNet (top@1 acc.)} & {50} & {78.96} & {0.94}  \\
{MobileNetV2} & {ImageNet (top@1 acc.)} & {40} & {70.29} & {1.51}  \\
{MobileNetV1-SSD} & {VOC07 (mAP)} & {50} & {71.53} & {0.98}  \\
{TinyYOLOv2} & {COCO (AP)} & {50} & {28.29} & {0.83}  \\
{NCF} & {MovieLens 20M (hit ratio)} & {70} & {64.67} & {0.93}  \\
{BERT-base} & {MRPC (acc.)} & {50} & {82.50} & {0.63}  \\
\bottomrule
\end{tabular*}
\end{table*}

\begin{table*}
\caption{Accuracy of sparse computer vision models obtained with layer-wise fine-tuning on different input data. Note that accuracy levels are very similar when fine-tuning on original validation or training datasets, suggesting the absence of overfitting during layer-wise fine-tuning. "FractalDB-1k(c)" denotes the colored FractalDB-1k dataset.}
\label{tab:tuning_data_results}
\begin{tabularx}{\linewidth}{lXXXXX}
\toprule
{Model (sparsity rate, dataset/acc. metric)} & {Orig. model acc.} & {Val. data} & {Training data} & {FractalDB 1k(c)} & {White noise}\\
\midrule

{ResNet18 (50\%; ImageNet top@1)} & {69.75} & {68.94} & {68.92} & {68.27} & {66.90} \\
{ResNet50 (50\%; ImageNet top@1)} & {76.13} & {75.51} & {75.57} & {74.50} & {73.89} \\
{MobileNetV2 (40\%; ImageNet top@1) }& {71.81} & {70.04} & {70.12} & {68.94} & {66.84} \\
{MobileNetV1-SSD (50\%; VOC07 mAP)} & {72.51} & {71.37} & {71.53} & {71.13} & {69.52} \\
{TinyYOLOv2 (50\%; COCO AP)} & {29.12} & {28.06} & {28.29} & {28.18} & {27.10} \\
{RetinaFace-ResNet50 (50\%; WIDER FACE mAP)} & {87.29} & {87.40} & {87.41} & {87.45} & {86.87} \\
\bottomrule
\end{tabularx}
\end{table*}

\section{Related work}

Neural network weight pruning is a technique used to produce lightweight models by removing (zeroing out) a certain percentage of unimportant weights. In this work, we focus on unstructured pruning (weight sparsification) whereby no structural constraints on the sparsity pattern are imposed and a subset of weights determined to have the lowest importance score values is removed regardless of position in weight tensors. Various definitions of weight importance functions have been proposed in the literature \cite{gale2019state, zhu2017prune}, the simplest baseline being magnitude-based weight pruning, as well as various heuristics to determine per-layer sparsity rates \cite{lee2020deeper, he2018amc}. Magnitude-based sparsification via a global importance threshold was found to be a strong baseline in the compression-aware training regime for a range of models \cite{gale2019state, singh2020woodfisher}. These weight pruning approaches typically imply compression-aware model re-training, which means existing access to the training code, the training dataset and appropriate compute resources. Other DNN compression techniques, such as 8-bit quantization, however, have been successfully applied in a less restrictive setting -- in the post-training or data-free regimes \cite{nagel2019data}. The post-training compression regime is favorable from a practical perspective, since model compression could be ultimately implemented via a single API call rather than via the modification of the original model training code. There, however, have been few attempts to implement post-training or data-free weight pruning of DNNs, primarily due to the large accuracy drop incurred during sparsification \cite{horton2020layer, srinivas2015data}. Recently, there have also been developed layer-wise gradient optimization-based methods for post-training compression \cite{hubara2020improving, li2021brecq, nagel2020up, horton2020layer} with applications to low-bitwidth quantization and weight pruning. These methods are promising because they allow restoring compressed model accuracy in the post-training setting in many cases. Nevertheless accuracy degradation was still found to be significant for sparsity rates above 40\%. In this work, we propose a post-training sparsification recipe that allows insignificant accuracy drops on a range of DNN models at sparsity rates of 50\% and higher. We also suggest a straightforward and fast extension of the method for the data-free compression of computer vision models, using synthetic fractal image data, that allows getting state-of-the-art accuracy on a range of natural image datasets.

\section{Post-training sparsity pipeline}

\begin{figure}
\includegraphics[width=\columnwidth]{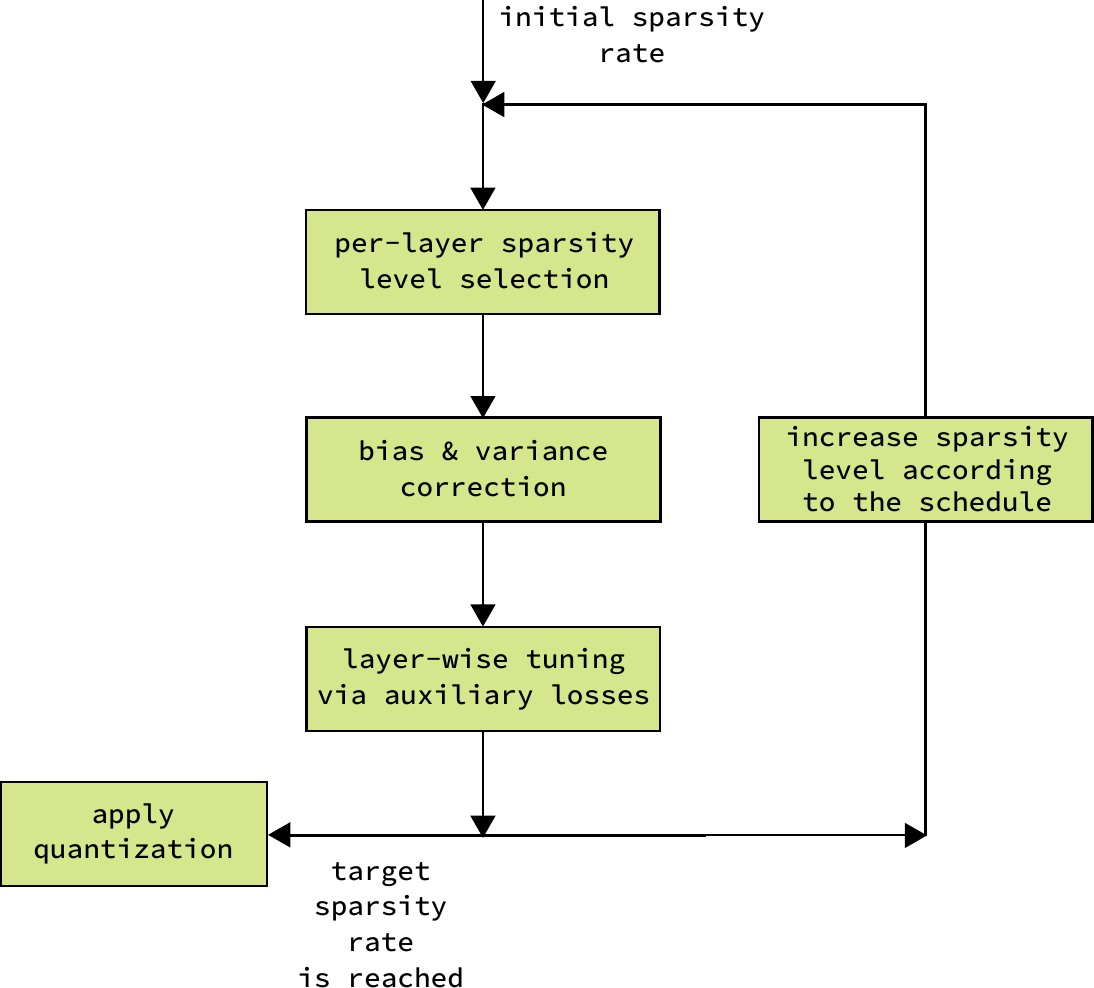}
\caption{Flowchart of the proposed post-training sparsity pipeline. The process begins with a small initial global sparsity value, the model is fine-tuned in a layer-wise manner, the sparsity level is increased and the fine-tuning is repeated. This iterative process is carried out until the target sparsity level or the maximal allowed accuracy drop is reached (either of these parameters is set in advance).}
\label{fig:pot_pipeline}
\end{figure}

\begin{figure}
    \centering
    \includegraphics[width=1.0\columnwidth]{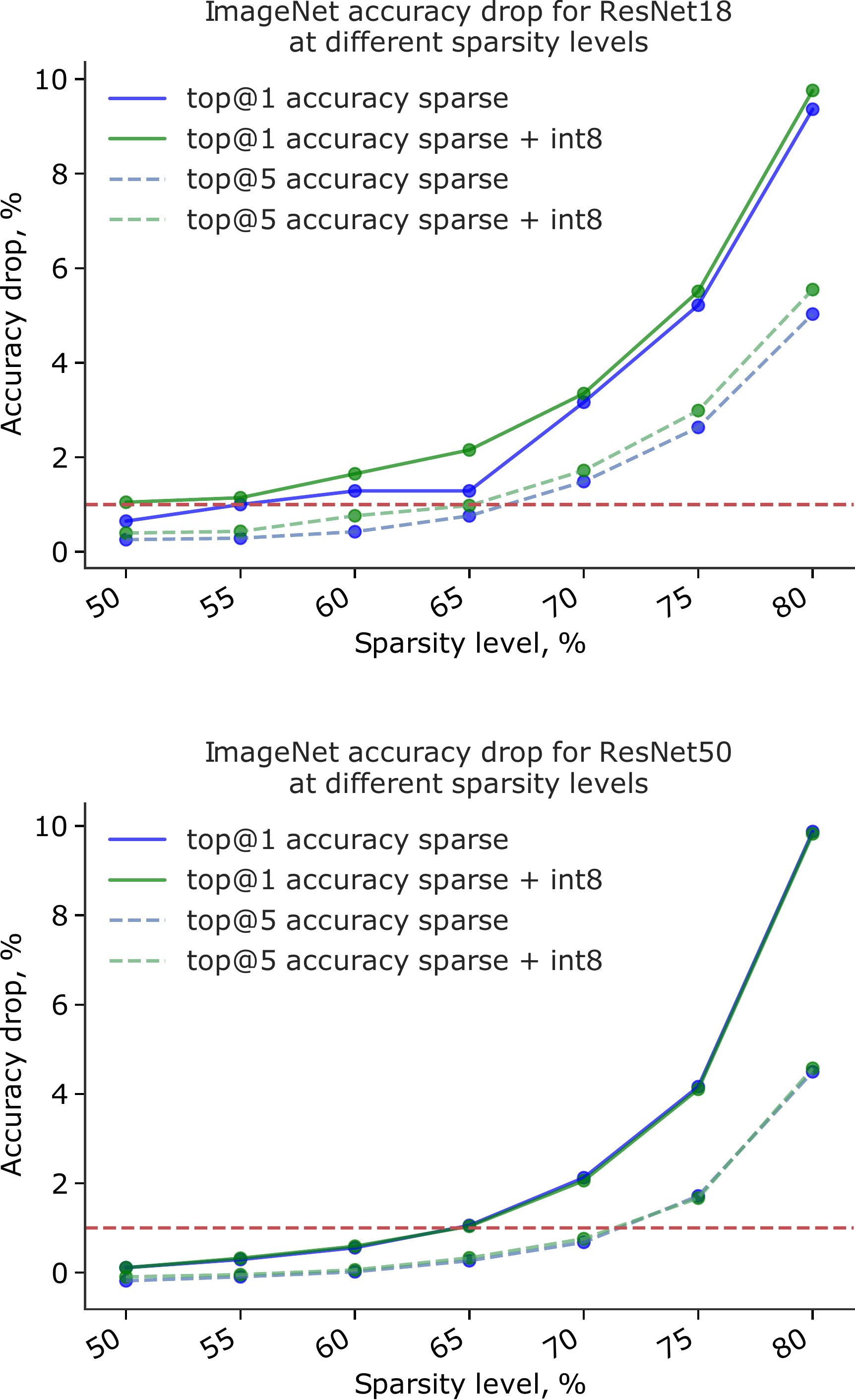}
    \caption{Accuracy drop/sparsity rate curves for a ResNet18 (top) and a ResNet50 (bottom) model obtained with our post-training pruning and quantization pipeline. The horizontal dashed red line indicates the level of 1\% absolute accuracy drop. Note that post-training quantization of the pruned model does not lead to a huge accuracy drop increase for both models at different sparsity rate levels.}
    \label{fig:resnet_sparse_results}
\end{figure}

The proposed post-training sparsity pipeline consists of three basic steps: (i) layer-wise sparsity rate selection given a global sparsity constraint, (ii) bias \& variance correction steps, and (iii) layer-wise fine-tuning using auxiliary knowledge distillation losses. We introduce a progressively increasing sparsity schedule for each layer whereby these three steps are performed iteratively and the global sparsity rate in the model is increased on each iteration (see the flowchart in Fig. \ref{fig:pot_pipeline}). The global sparsity rate for the model on the $t^{th}$ iteration of the pipeline is determined via the following polynomial (cubic) sparsity schedule \cite{zhu2017prune}: 
\begin{equation*}
    s_t = s_f + (s_i - s_f)\left(1 - \frac{t}{T}\right)^3
\end{equation*}
where $s_i$ and $s_f$ are the initial and final global sparsity rates of the model, respectively, and $T$ is the total number of iterations of the pipeline. After the original floating-point precision model with the target global sparsity rate is obtained, the standard procedure of post-training quantization is performed to prepare the model to be executed in 8-bit precision. We found that performing post-training quantization on the pruned model does not incur significant accuracy degradation compared to the original-precision sparse model (see Fig. \ref{fig:resnet_sparse_results} for results on ResNet18/50 on ImageNet), probably due to reduced quantization noise of sparse weight matrices. We are using the following quantization configuration throughout the paper: symmetric per-tensor quantization of activations (except for specific per-channel cases like e.g. depthwise convolutions) and symmetric per-channel quantization of weights. We further provide details on all the steps performed on every iteration of the pruning pipeline in the corresponding sections below.

\subsubsection*{Layer-wise sparsity rate selection procedure}

\begin{table}
\caption{Impact of per-layer sparsity selection criteria on a pre-trained ResNet50 model on ImageNet with 50\% of weights pruned.}
\begin{tabularx}{\linewidth}{lXX}
\toprule
Sparsity selection criterion & BN-fusing & Top@1 accuracy, \%\\
\midrule
ResNet50 50\% sparsity \\ \hline
{Original model} & {} & {76.13} \\ \hline
{Magnitude} & {Yes} & {0.3814} \\ 
{L2-normalized magnitude} & {Yes} & {\bf{72.544}} \\
{LAMP} & {Yes} & {72.328} \\ \hline
{Magnitude} & {No} & {\bf{72.831}} \\ 
{L2-normalized magnitude} & {No} & {72.244} \\ \hline
ResNet18 50\% sparsity \\ \hline
{Original model} & {} & {69.75} \\ \hline
{Magnitude} & {Yes} & {0.418} \\ 
{L2-normalized magnitude} & {Yes} & {64.856} \\ 
{LAMP} & {Yes} & {\bf{64.866}} \\ \hline
MobileNetV2 30\% sparsity \\ 
(with bias correction) \\ \hline
{Original model} & {} & {71.81} \\ \hline
{Magnitude} & {Yes} & {18.386} \\ 
{L2-normalized magnitude} & {Yes} & {\bf{69.528}} \\ 
{LAMP} & {Yes} & {69.524} \\
\bottomrule
\end{tabularx}
\label{fig:per_layer_table}
\end{table}

\begin{figure*}
    \centering
    \includegraphics[width=0.9\textwidth]{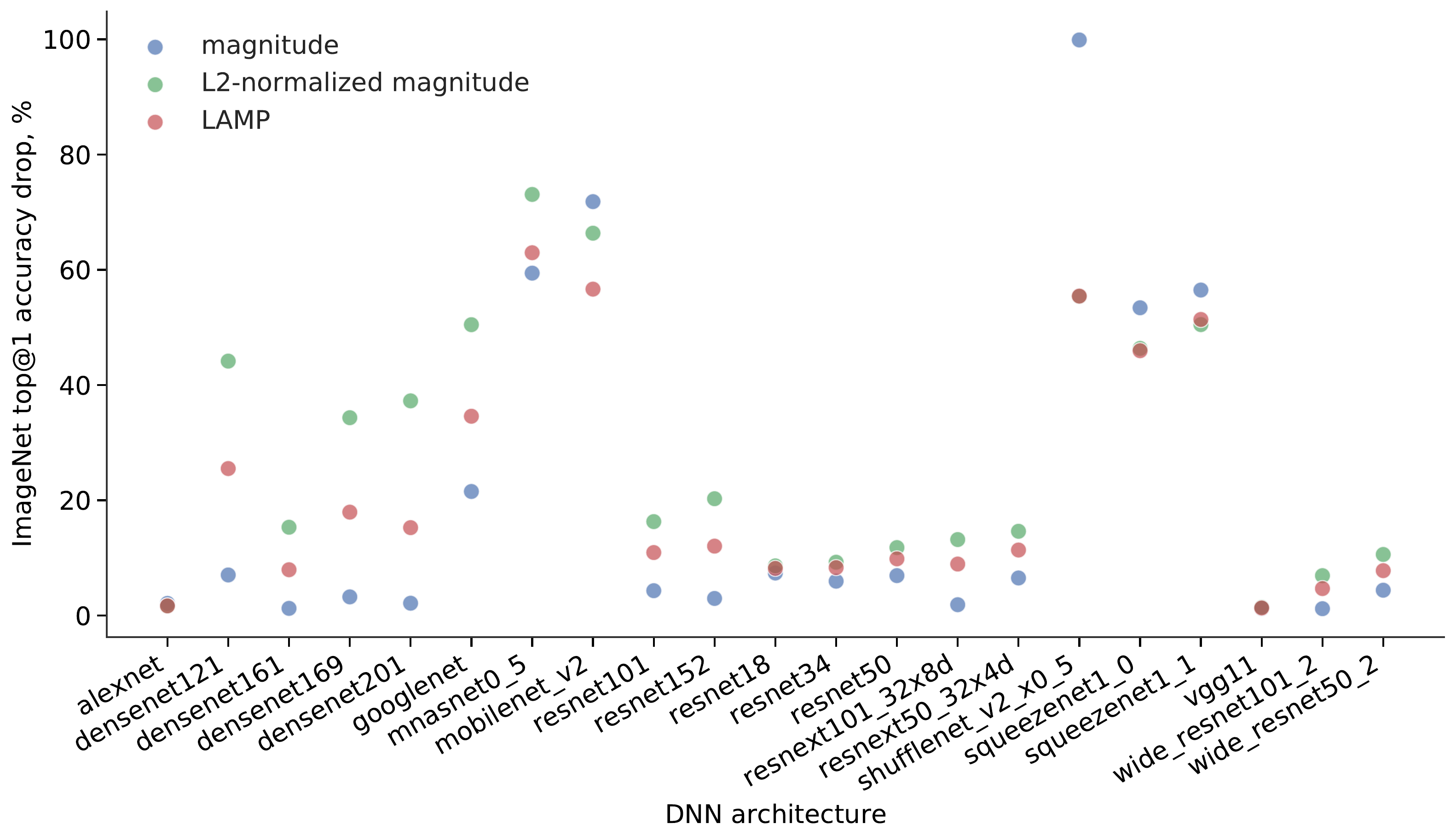}
    \caption{Absolute top@1 accuracy drops for a range of models from the torchvision package pruned with a 50\% global sparsity rate depending on the per-layer compression level selection criterion. BatchNorm fusing was not performed in the networks. Pruning is done in the post-training regime (without any fine-tuning), BatchNorm adaptation is performed after weight pruning. Global magnitude criterion is optimal in most cases in this setting except for lightweight models such as MobileNets, SqueezeNets and ShuffleNets. BatchNorm adaptation is a procedure analogous to bias correction whereby the BN statistics are recollected after the model has been compressed \cite{li2020eagleeye}. Accuracy drops are measured relative to the original pre-trained weights using PyTorch on a GPU.}
    \label{fig:layerwise_scores_torchvision}
\end{figure*}


The problem of selecting an optimal (in terms of model accuracy) layer-wise sparsity rate configuration given a certain global sparsity constraint is a widely discussed problem in the literature \cite{lee2020deeper}. The proposed approaches range from simple heuristics (e.g. pruning uniformly except for the first and the last layers in the network \cite{gale2019state}) to making per-layer sparsity rates learnable \cite{kusupati2020soft} or searching for the best configuration via global non-gradient optimization or reinforcement learning \cite{he2018amc}. The heuristic approaches also include finding a global threshold for weight importance scores and pruning all the weights with importances below this threshold. This naturally leads to a non-uniform pattern of per-layer sparsity rates. The importance score function in this case might be the absolute weight magnitude or a normalized version thereof (like e.g. the LAMP score \cite{lee2020deeper}). We compared several variations of importance score functions in the global threshold approach, namely (i) the absolute weight magnitude, (ii) the absolute weight magnitude normalized by the L2 norm of the corresponding layer, (iii) the LAMP score (Table \ref{fig:per_layer_table}). We initally observed that the global magnitude criterion led to much worse accuracy comprared to the normalized criteria (Table \ref{fig:per_layer_table}). This effect was found to be caused by the fusing of the BatchNorm layers into preceding convolutions, which was performed in the model prior to compression. BatchNorm fusing resulted in different layer-wise weight scales compared to the original model, an effect easily counteracted by per-layer normalization of weight magnitudes. In the case where the normalization layers were not fused into convolutions, however, we found that the vanilla global magnitude criterion performed the best compared to LAMP and L2-normalized magnitude (Table \ref{fig:per_layer_table} and Figure \ref{fig:layerwise_scores_torchvision}). We further assumed that the fusing could generally occur prior to model compression and the original normalization layer parameters might be unknown, hence we picked the per-layer L2-normalized magnitude criterion as our sparsity rate selection heuristic. It performed better than LAMP in our post-training scenario on most of the models with BatchNorm fusing. The weight importance criterion for the $i^{th}$ weight in the $l^{th}$ layer $w_i^l$ we use in our pipeline thus reads
\begin{equation*}
    I(w^l_i) = \frac{|w_i^l|}{\sqrt{\sum_{j \in l} |w_j^l|^2 }}
\end{equation*}
We pool the importance scores from all the layers and find the threshold value corresponding to the set sparsity rate. The weights with importance values below the threshold are pruned. 

\subsubsection*{Weight and activation bias correction}

Once the layer-wise pruning rates have been determined, the weights are zeroed out based on the intra-layer absolute magnitudes. This pruning operation distorts the weight distribution, introducing bias and scale shifts. It is beneficial to carry out a bias correction procedure on the weights in order to restore the original mean and variance values in all of the convolutional layer filters and fully-connected layer weight matrices \cite{banner2018post}. We perform the following affine transformation on all of the pruned weight tensors in a per-channel/per-feature fashion:
\begin{equation*}
    W^s_{corr} = \lambda W^s + E(W_{dense}) - E(\lambda W^s)
\end{equation*}
\begin{equation*}
    \lambda = \frac{\sigma(W_{dense})}{\sigma(W^s) + \epsilon}
\end{equation*}
where $W^s_{corr}$ is the weight tensor after the correction procedure, and $W_s$ and $W_{dense}$ are the weight tensors of sparse and original dense models, respectively, and $E$ and $\sigma$ are the mean and standard deviation operators, $\epsilon=10^{-9}$ is a small constant added for numerical stability. The resulting sparse weight tensor has the same mean and variance values as original dense model weights for each output kernel/feature, since this correction is applied to every output feature independently. 
Output activations at each pruned layer are also suffering from a bias introduced by the zeroed weights, which can be compensated by altering the bias parameters of the convolutional and fully-connected layers. Nagel et al. \cite{nagel2019data} proposed to perform this operation to mitigate biases introduced by quantization in an iterative fashion, correcting the first layer and then calculating the bias shift factors for the second layer using this corrected model. We found that a one-shot version of the bias correction procedure was sufficient for post-training sparsity, whereby we perform a forward pass of the original model and calculate the input activation tensors $X_{dense}$ for each layer. The corrected bias parameters are then determined as
\begin{multline*}
    b_{corr} = b_{dense} + E(f(W_{dense}, X_{dense})) - \\ E(f(W^s_{corr}, X_{dense}))
\end{multline*}
where $f(W, X)$ is the convolutional or matrix-multiply operation of the layer acting on inputs $X$ with weights $W$, $b_{dense}$ are the original bias values in the layer, $X_{dense}$ is the set of input activation tensors for the corresponding layer in the original dense model. In other words, we are using the input tensors from the original model to calculate bias shifts, not from the iteratively corrected compressed model. We found no significant difference in the resulting accuracy between the two approaches, with the one-shot one being faster since it requires a single forward pass of the model. Results of the weight \& activation bias correction procedures are shown in Table \ref{tab:bias_corr} for a ResNet18 model at 50\% sparsity rate. Both procedures cumulatively improve the pruned model accuracy and top@1 accuracy drops are not exceeding several percent for many ImageNet models at the sparsity rate of 50\% just after layer-wise sparsity selection and bias correction. Accuracy can be further improved by local layer-wise fine-tuning using auxiliary knowledge-distillation losses, which is described in more detail below. 

\subsubsection*{Local layer-wise fine-tuning with auxiliary losses}

\begin{figure}
\includegraphics[width=\columnwidth]{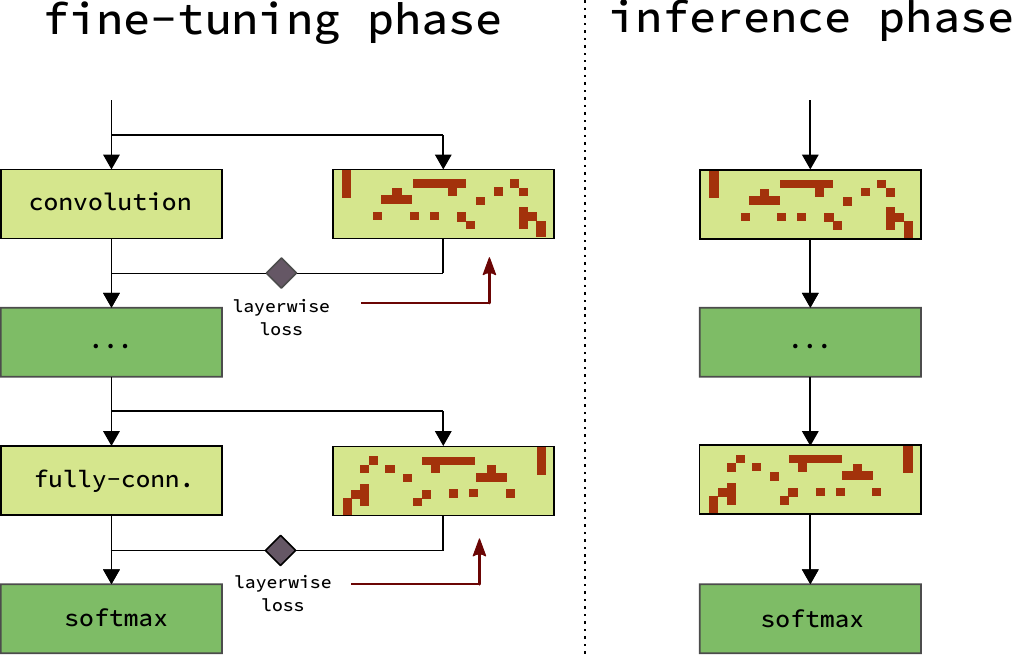}
\caption{Schematic description of the layerwise fine-tuning approach for post-training sparse model calibration. Input and dense model output tensors are pre-computed and stored in memory for each tuned layer. The red arrows depict the local flow of gradients during weight and bias optimization. Red pixels indicate sparsity masks in the compressed layers (sparsity levels are individually selected for each layer).}
\label{fig:finetuning_pipeline}
\end{figure}

Gradient-based fine-tuning with auxiliary loss functions was previously successfully applied for post-training quantization \cite{nagel2020up, li2021brecq, hubara2020improving} and, to a smaller extent, to post-training weight pruning \cite{horton2020layer} and even filter pruning in convolutional networks \cite{guan2019wootz}. In this work, we follow a similar approach whereby we define a local knowledge distillation loss for every pruned layer (see Figure \ref{fig:finetuning_pipeline}). These losses serve as a measure of how close the output activation feature maps of the original (unpruned) and pruned layers are. Suppose the pruned layer weights and biases are $W^s$ and $b^s$, then the knowledge distillation mean-squared error loss for that layer is defined as 
\begin{multline*}
    L = \sum_{i \in batch} (Y^i_{dense} - f(W^sM^s, X^i_{dense}) - b^s)^2\\
    Y^i_{dense} = f(W_{dense}, X^i_{dense})
\end{multline*}
where $f(X, W)$ is the convolutional/matrix-multiply operation represented by the layer with weights $W$ acting on inputs $X$. The input tensors $X_{dense}$ used to calculate the output activations above are constructed by running a forward pass of the original, unpruned model. $M^s$ is the binary mask layer which is equal to one if the corresponding weight is not pruned and zero otherwise. We fix the sparse binary mask and run a gradient descent of the loss functions defined above for each layer independently to find the optimal weights and biases $W^s$, $b^s$.

\subsubsection*{Ablation study}

\textit{Layer-wise fine-tuning settings.} We run several ablation experiments to establish the best optimization settings for the layer-wise fine-tuning procedure, since it is inherently different from full model training via backpropagation. We used a batch size of 50 samples in our experiments, and found the optimal learning rate values across different models to be $10^{-5}$ for weights and $10^{-4}$ for bias parameters. We found that using the Adam optimizer outperforms alternatives, such as SGD with momentum or Adadelta, and techniques for better generalization in the vicinity of local minima, like Lookahead \cite{zhang2019lookahead} and Stochastic Weight Averaging \cite{izmailov2018averaging} were also not found to be beneficial in the layer-wise fine-tuning case. The MSE loss function also was found to be a better choice than e.g. L1 loss, Huber loss or cosine similarity between feature maps. We did not observe significant over-fitting present in layer-wise optimization (we discuss this phenomenon in more details below) and in particular we found that even low values of weight decay/L2 weight regularization strength such as $10^{-6}$ could hurt the resulting model accuracy (see Table \ref{tab:weigth_decay_tuning}). \textit{Thus, we set the weight decay strength to 0 in all our experiments.} Increasing model sparsity rate using a cubic schedule throughout the pruning pipeline also turned out to improve accuracy for most models compared to the constant sparsity baseline (Table \ref{tab:schedule_tuning}). Overall, we were able to prune and quantize a wide range of models with resulting sparsity rates ranging from 40\% to 70\% and an absolute accuracy drop not exceeding or close to 1\% with our layer-wise fine-tuning recipe using images from the respective models' training datasets (Table \ref{table:sparse_int8_results} and Figure \ref{fig:resnet_sparse_results}). 

\begin{table}
\caption{Impact of bias \& variance correction for weights and activations on a ResNet18 model on ImageNet with 50\% of the weights pruned.}
\begin{tabularx}{\linewidth}{lXX}
\toprule
ResNet18 50\% sparse & Top@1 accuracy & Top@5 accuracy \\
\midrule
L2-normalized magnitude & 64.856 & 86.126 \\
L2-normalized magnitude \\ (+ act. bias correction) & 66.852 & 87.438 \\
L2-normalized magnitude \\ (+ act. \& weight bias correction) & \bf{67.46} & \bf{87.794} \\
\bottomrule
\end{tabularx}
\label{tab:bias_corr}
\end{table}

\begin{table}
\caption{Impact of L2 weight regularization on the fine-tuned model accuracy for a ResNet18 model at 50\% sparsity on ImageNet. Even small weight decay values result in significant accuracy loss compared to the baseline with no regularization in place.}
\label{tab:weigth_decay_tuning}
\begin{tabularx}{\linewidth}{cc}
\toprule
L2-reg. strength & Top@1 / Top@5 accuracy, \%\\
\midrule
{$\lambda =$ 0.0} & {\bf{68.97 / 88.77}} \\ 
{$\lambda =$ 1e-6} & {67.78 / 87.96} \\
{$\lambda =$ 1e-5} & {38.14 / 64.49} \\
\bottomrule
\end{tabularx}
\end{table}

\textit{Source of input data used for fine-tuning.} Throughout experiments, we noticed that the set of input samples to be used for fine-tuning does not have to be necessarily large (we used a pool of randomly selected several hundred samples in our experiments) and can come either from the training or the validation dataset, with no significant accuracy difference between the two (see Table \ref{tab:tuning_data_results}). These results suggest that this fine-tuning regime is not as prone to over-fitting compared to full model training, a fact that was also previously reported for layer-wise tuning of a quantized model \cite{hubara2020improving}. We observed no difference between fine-tuning on a batch of training or validation samples not only for the ImageNet dataset but also for object detection models trained on the Pascal VOC, COCO and WIDER FACE datasets. This lack of over-fitting is not surprising since no annotation is used during fine-tuning and all the layers are optimized independently, which reduces the amount of tuned parameters per single optimization problem. The amount of supervision signal is also high because the difference between whole activation tensors produced by a set of input samples is used as a loss function.
We further verified whether we could utilize arbitrary input data for bias correction and layer-wise fine-tuning, not related to the original dataset that the model has been trained and tested on. The intuition behind this is that once the output activation feature maps produced by these arbitrary data are similar to the ones produced by running model inference on its original dataset, the layer-wise fine-tuning procedure could produce a model that is sufficiently accurate on the validation data. In particular, we tested several computer vision models trained on different datasets (ImageNet, Pascal VOC, COCO, WIDER FACE; see Table \ref{tab:tuning_data_results}); the models were pruned using our post-training pipeline, but the input images used to calculate activation statistics and feature maps for fine-tuning consisted of synthetically generated white noise (each pixel value in every color channel is independently sampled from a uniform distribution from 0 to 255). We observed certain accuracy degradation when fine-tuning on white noise images compared to tuning on original data, but typically not exceeding several percent. We further tested whether these results could be improved  by using synthetic images producing activation distributions closer to those generated by natural images in corresponding datasets. We took images from the FractalDB-1k dataset \cite{kataoka2020pre}, which is comprised of automatically-generated grayscale images of fractals. These images and their generated annotation were used to pre-train strong backbones for computer vision, including Vision Transformers \cite{kataoka2020pre, nakashima2021can}. We found that using these fractal images as input samples to computer vision models during post-training weight pruning significantly improves the resulting model accuracy compared to the white noise baseline (Table \ref{tab:tuning_data_results}). We took the original 512x512 images from FractalDB-1k, randomly colored them by performing random shift-scale operations on the color channels and used the same pre-processing strategy as for the original datasets that the models were trained on. Overall, we were able to achieve an accuracy degradation of absolute 1-3\% at 50\% sparsity rates in the data-free pruning regime by using synthetically-generated fractal images as model inputs. The results generalized beyond ImageNet to other natural image datasets like Pascal VOC, COCO and WIDER FACE. Random colorization of FractalDB images consistently yielded better accuracy compared to using the (original) grayscale images (Table \ref{tab:colored_vs_grayscale}). The proposed data-free pruning approach leads to \textit{better accuracy values compared to the existing state-of-the-art} \cite{horton2020layer} and is also fast and less restrictive since it does not include a resource-consuming data distillation process that relies on backpropagation through the model graph.


\begin{table}
\caption{Impact of the sparsity schedule vs. constant sparsity throughout fine-tuning. All metrics are reported at 50\% sparsity rates. The cubic schedule was initialized at 10\% sparsity rate which was increased in 10 iterations. The same number of optimizer steps was used for both fine-tuning modes.}
\label{tab:schedule_tuning}
\begin{tabularx}{\linewidth}{lXX}
\toprule
{Model} & {Top@1 accuracy w/o schedule} & {Top@1 accuracy with schedule}\\
\midrule
{ResNet18} & {68.76} & {\bf{68.91}} \\
{ResNet50} & {75.43} & {\bf{75.60}} \\
{GoogleNetV4} & {\bf{79.37}} & {78.10} \\
\bottomrule
\end{tabularx}
\end{table}

\begin{table*}
\caption{Accuracy of models pruned in the post-training regime using different types of synthetic data: randomly-colored FractalDB1k images, (original) grayscale FractalDB-1k, and generated white noise images. Note that colorization of FractalDB images leads to increased resulting accuracy in most models.}
\label{tab:colored_vs_grayscale}
\begin{tabularx}{\linewidth}{lXXX}
\toprule
{Model (sparsity rate, dataset/acc. metric)} & {FractalDB-1k(c)} & {FractalDB-1k} & {White noise}\\
\midrule
{ResNet18 (50\%; ImageNet top@1)} & {\bf{68.27}} & {67.94} & {66.90} \\
{ResNet50 (50\%; ImageNet top@1)} & {\bf{74.50}} & {74.46} & {73.89} \\
{MobileNetV2 (40\%; ImageNet top@1) } & {\bf{68.94}} & {68.47} & {66.84} \\
{MobileNetV1-SSD (50\%; VOC07 mAP)} & {\bf{71.13}} & {70.79} & {69.52} \\
{TinyYOLOv2 (50\%; COCO AP)} & {28.18} & {\bf{28.28}} & {27.10} \\
\end{tabularx}
\end{table*}



\section{Conclusion}

In this work, we have presented a novel post-training pruning recipe for deep neural networks that allows zeroing out a significant proportion of model weights without significant accuracy drops. We demonstrated efficiency of the proposed pipeline on ImageNet models, object detection models on Pascal VOC and COCO datasets as well as deep NLP and recommendation models. We proposed a data-free formulation of the method by using synthetic fractal images to compress computer vision models, which led to state-of-the-art results in data-free weight pruning. We demonstrated that the proposed pruning method can also be safely combined with post-training quantizaton, further increasing its applicability in production settings.

{\small
\bibliographystyle{ieee}
\bibliography{egbib}

\begin{thebibliography}{10}\itemsep=-1pt

\bibitem{banner2018post}
R.~Banner, Y.~Nahshan, E.~Hoffer, and D.~Soudry.
\newblock Post-training 4-bit quantization of convolution networks for
  rapid-deployment.
\newblock {\em arXiv preprint arXiv:1810.05723}, 2018.

\bibitem{blalock2020state}
D.~Blalock, J.~J.~G. Ortiz, J.~Frankle, and J.~Guttag.
\newblock What is the state of neural network pruning?
\newblock {\em arXiv preprint arXiv:2003.03033}, 2020.

\bibitem{elsen2020fast}
E.~Elsen, M.~Dukhan, T.~Gale, and K.~Simonyan.
\newblock Fast sparse convnets.
\newblock In {\em Proceedings of the IEEE/CVF Conference on Computer Vision and
  Pattern Recognition}, pages 14629--14638, 2020.

\bibitem{gale2019state}
T.~Gale, E.~Elsen, and S.~Hooker.
\newblock The state of sparsity in deep neural networks.
\newblock {\em arXiv preprint arXiv:1902.09574}, 2019.

\bibitem{guan2019wootz}
H.~Guan, X.~Shen, and S.-H. Lim.
\newblock Wootz: A compiler-based framework for fast cnn pruning via
  composability.
\newblock In {\em Proceedings of the 40th ACM SIGPLAN Conference on Programming
  Language Design and Implementation}, pages 717--730, 2019.

\bibitem{guo2020accelerating}
C.~Guo, B.~Y. Hsueh, J.~Leng, Y.~Qiu, Y.~Guan, Z.~Wang, X.~Jia, X.~Li, M.~Guo,
  and Y.~Zhu.
\newblock Accelerating sparse dnn models without hardware-support via tile-wise
  sparsity.
\newblock {\em arXiv preprint arXiv:2008.13006}, 2020.

\bibitem{he2018amc}
Y.~He, J.~Lin, Z.~Liu, H.~Wang, L.-J. Li, and S.~Han.
\newblock Amc: Automl for model compression and acceleration on mobile devices.
\newblock In {\em Proceedings of the European Conference on Computer Vision
  (ECCV)}, pages 784--800, 2018.

\bibitem{horton2020layer}
M.~Horton, Y.~Jin, A.~Farhadi, and M.~Rastegari.
\newblock Layer-wise data-free cnn compression.
\newblock {\em arXiv preprint arXiv:2011.09058}, 2020.

\bibitem{hubara2020improving}
I.~Hubara, Y.~Nahshan, Y.~Hanani, R.~Banner, and D.~Soudry.
\newblock Improving post training neural quantization: Layer-wise calibration
  and integer programming.
\newblock {\em arXiv preprint arXiv:2006.10518}, 2020.

\bibitem{izmailov2018averaging}
P.~Izmailov, D.~Podoprikhin, T.~Garipov, D.~Vetrov, and A.~G. Wilson.
\newblock Averaging weights leads to wider optima and better generalization.
\newblock {\em arXiv preprint arXiv:1803.05407}, 2018.

\bibitem{kataoka2020pre}
H.~Kataoka, K.~Okayasu, A.~Matsumoto, E.~Yamagata, R.~Yamada, N.~Inoue,
  A.~Nakamura, and Y.~Satoh.
\newblock Pre-training without natural images.
\newblock In {\em Proceedings of the Asian Conference on Computer Vision},
  2020.

\bibitem{kusupati2020soft}
A.~Kusupati, V.~Ramanujan, R.~Somani, M.~Wortsman, P.~Jain, S.~Kakade, and
  A.~Farhadi.
\newblock Soft threshold weight reparameterization for learnable sparsity.
\newblock In {\em International Conference on Machine Learning}, pages
  5544--5555. PMLR, 2020.

\bibitem{lee2020deeper}
J.~Lee, S.~Park, S.~Mo, S.~Ahn, and J.~Shin.
\newblock A deeper look at the layerwise sparsity of magnitude-based pruning.
\newblock {\em arXiv preprint arXiv:2010.07611}, 2020.

\bibitem{li2020eagleeye}
B.~Li, B.~Wu, J.~Su, and G.~Wang.
\newblock Eagleeye: Fast sub-net evaluation for efficient neural network
  pruning.
\newblock In {\em European Conference on Computer Vision}, pages 639--654.
  Springer, 2020.

\bibitem{li2021brecq}
Y.~Li, R.~Gong, X.~Tan, Y.~Yang, P.~Hu, Q.~Zhang, F.~Yu, W.~Wang, and S.~Gu.
\newblock Brecq: Pushing the limit of post-training quantization by block
  reconstruction.
\newblock {\em arXiv preprint arXiv:2102.05426}, 2021.

\bibitem{liu2020layerwise}
X.~Liu, W.~Li, J.~Huo, L.~Yao, and Y.~Gao.
\newblock Layerwise sparse coding for pruned deep neural networks with extreme
  compression ratio.
\newblock In {\em Proceedings of the AAAI Conference on Artificial
  Intelligence}, volume~34, pages 4900--4907, 2020.

\bibitem{nagel2020up}
M.~Nagel, R.~A. Amjad, M.~Van~Baalen, C.~Louizos, and T.~Blankevoort.
\newblock Up or down? adaptive rounding for post-training quantization.
\newblock In {\em International Conference on Machine Learning}, pages
  7197--7206. PMLR, 2020.

\bibitem{nagel2019data}
M.~Nagel, M.~v. Baalen, T.~Blankevoort, and M.~Welling.
\newblock Data-free quantization through weight equalization and bias
  correction.
\newblock In {\em Proceedings of the IEEE/CVF International Conference on
  Computer Vision}, pages 1325--1334, 2019.

\bibitem{nakashima2021can}
K.~Nakashima, H.~Kataoka, A.~Matsumoto, K.~Iwata, and N.~Inoue.
\newblock Can vision transformers learn without natural images?
\newblock {\em arXiv preprint arXiv:2103.13023}, 2021.

\bibitem{singh2020woodfisher}
S.~P. Singh and D.~Alistarh.
\newblock Woodfisher: Efficient second-order approximation for neural network
  compression.
\newblock {\em Advances in Neural Information Processing Systems}, 33, 2020.

\bibitem{srinivas2015data}
S.~Srinivas and R.~V. Babu.
\newblock Data-free parameter pruning for deep neural networks.
\newblock {\em arXiv preprint arXiv:1507.06149}, 2015.

\bibitem{zhang2019lookahead}
M.~R. Zhang, J.~Lucas, G.~Hinton, and J.~Ba.
\newblock Lookahead optimizer: k steps forward, 1 step back.
\newblock {\em arXiv preprint arXiv:1907.08610}, 2019.

\bibitem{zhu2017prune}
M.~Zhu and S.~Gupta.
\newblock To prune, or not to prune: exploring the efficacy of pruning for
  model compression.
\newblock {\em arXiv preprint arXiv:1710.01878}, 2017.

\end{thebibliography}
}

\end{document}